\documentclass[10pt,twocolumn,letterpaper]{article}

\usepackage[pagenumbers]{wacv}
\usepackage{graphicx}
\usepackage{amsmath}
\usepackage{amssymb}
\usepackage{booktabs}
\usepackage{multirow}
\usepackage{makecell}
\usepackage[pagebackref,breaklinks,colorlinks]{hyperref}

\usepackage[capitalize]{cleveref}
\crefname{section}{Sec.}{Secs.}
\Crefname{section}{Section}{Sections}
\Crefname{table}{Table}{Tables}
\crefname{table}{Tab.}{Tabs.}

\begin{document}
\title{Integrating Meshes and 3D Gaussians for Indoor Scene Reconstruction \\ with SAM Mask Guidance}

\author{{Jiyeop Kim \qquad Jongwoo Lim}\\
IPAI, Seoul National University\\
{\tt\small \{99edward,jongwoo.lim\}@snu.ac.kr}
}

\maketitle

\begin{figure*}[t]
    \centering
    \includegraphics[width=0.8\linewidth]{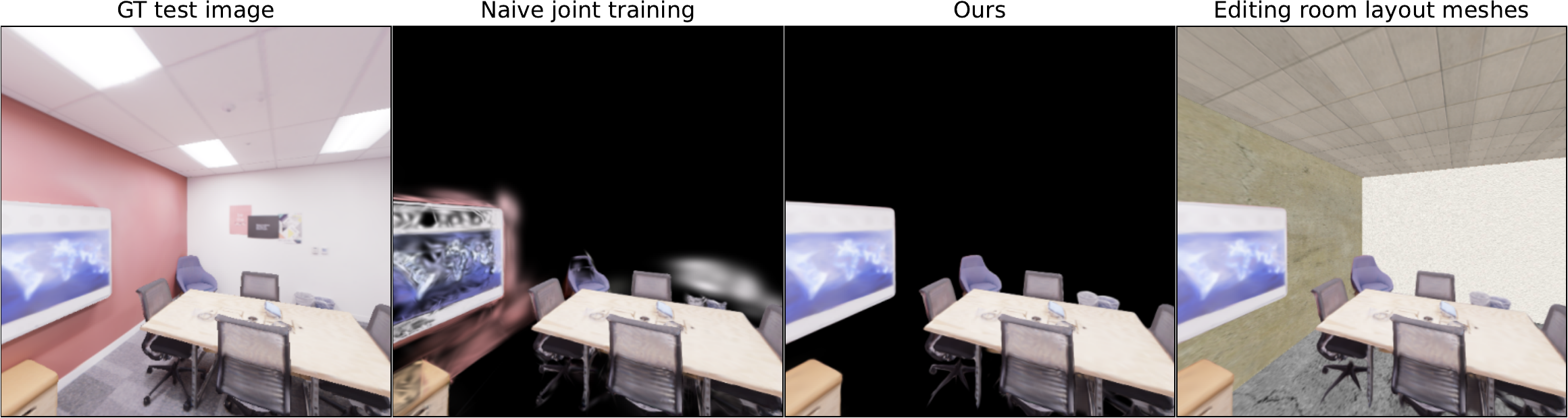}
    \caption{We propose a novel approach that integrates meshes into 3D Gaussian Splatting \cite{kerbl20233d} to represent the room layout of indoor scenes with meshes. Naively training 3D Gaussians and room layout meshes jointly can result in ambiguity issues, leading to a single instance being represented by both primitives. We address this issue by using SAM masks as a guide. Since the scene is represented as the sum of two primitives, we can selectively render or edit one of the primitives.}
    \label{fig:title}
\end{figure*}

\begin{abstract}
   We present a novel approach for 3D indoor scene reconstruction that combines 3D Gaussian Splatting (3DGS) with mesh representations. We use meshes for the room layout of the indoor scene, such as walls, ceilings, and floors, while employing 3D Gaussians for other objects. This hybrid approach leverages the strengths of both representations, offering enhanced flexibility and ease of editing.
   However, joint training of meshes and 3D Gaussians is challenging because it is not clear which primitive should affect which part of the rendered image. Objects close to the room layout often struggle during training, particularly when the room layout is textureless, which can lead to incorrect optimizations and unnecessary 3D Gaussians. 
   To overcome these challenges, we employ Segment Anything Model (SAM) to guide the selection of primitives. The SAM mask loss enforces each instance to be represented by either Gaussians or meshes, ensuring clear separation and stable training.
   Furthermore, we introduce an additional densification stage without resetting the opacity after the standard densification. This stage mitigates the degradation of image quality caused by a limited number of 3D Gaussians after the standard densification.
\end{abstract}

\section{Introduction}
\label{sec:intro}

Recently, 3D Gaussian Splatting (3DGS) \cite{kerbl20233d} has emerged for novel view synthesis and 3D scene reconstruction, offering real-time speeds with quality comparable to that of NeRF \cite{mildenhall2021nerf}. Unlike NeRF, which uses implicit representations, 3DGS reconstructs scenes using explicit 3D Gaussians. By sorting those 3D Gaussians by depth and alpha blending them sequentially from front to back, it can render images in real-time.

While 3D Gaussians serve as effective primitives for 3D scene reconstruction, they present several issues, such as floaters and unsmooth normals, and can also be inefficient; for instance, a simple plane composed of a few colors can be effectively represented with a small number of meshes, whereas representing it with 3D Gaussians requires a significantly larger number. Additionally, since the 3D Gaussians in a planar region do not directly interact with each other, the plane fails to maintain consistent normal. This is not only physically inaccurate, but also complicates applications such as mesh extraction.

We propose to use both 3D Gaussians and meshes for 3D indoor scene reconstruction. We model the walls, ceiling, and floor of a room with a few meshes, which are then jointly trained with 3D Gaussians. For texture mapping of the meshes, we use spherical harmonics (SH) instead of the commonly used RGB colors, which provides much greater rendering power and achieves a level of performance comparable to 3DGS. Our goal is to train meshes and 3D Gaussians so that the walls, ceiling, and floor of the scene are represented exclusively by meshes, while other objects are represented solely by 3D Gaussians. The trained layout meshes and 3D Gaussians can be rendered together or seperately, providing significant advantages in terms of editability and application.

Naively training room layout meshes and 3D Gaussians together can lead to several ambiguity problems as shown in \cref{fig:title}. Since both meshes and Gaussians contribute to the rendered color which is matched to the color of the ground truth (GT) image, it becomes unclear which component should affect the rendered image. Despite using multi-view images as training images, objects close to the room layout struggle with joint training, especially when the room layout is textureless. 

Another issue with joint training of 3D Gaussians and meshes is that any incorrect reconstruction by either the Gaussians or the room layout meshes can interrupt the optimization of the other. For example, if the learnable mesh representing the ceiling is incorrectly optimized, unnecessary 3D Gaussians are generated in front of it to match the GT images with the rendered images. Thus, naively training Gaussians and meshes together makes it challenging to achieve proper training for each as intended.

To address these issues, we employed Segment Anything Model (SAM) \cite{kirillov2023segment}, the 2D foundation model for image segmentation. SAM performs segmentation based on desired prompts for an input image and also can provide segmentation masks for all instances without prompts. Typically, after jointly training Gaussians and layout meshes, the opacity map from 3D Gaussians reveals high opacity for object parts and low opacity for room layout parts. Therefore, we used the SAM mask loss to enforce each instance to be represented either by Gaussians or the room layout meshes by ensuring the opacity values corresponding to each segmentation mask are close to 0 or 1. 

Since the SAM mask loss depends on the opacity map at each viewpoint, it conflicts with the densification process of standard Gaussian splatting, which periodically resets the opacity of 3D Gaussians. To avoid this, we apply the SAM mask loss after the densification is completed. However, as the number of 3D Gaussians becomes fixed after the densification, the rendered images become blurred, especially degrading the quality at object boundaries. To address this, we introduced an additional densification stage that does not reset the opacity. 

Our contributions are summarized as follows:
\begin{itemize}
    \item To the best of our knowledge, we are the first to reconstruct 3D scenes using two primitives: 3D Gaussians for objects and meshes with spherical harmonics for room layouts.
    \item To address the ambiguity problem arising from jointly training two primitives, we utilized masks obtained from the 2D foundation model, Segment Anything Model (SAM).
    \item To improve rendering quality, we introduce additional densification without opacity reset, applied after the standard densification. 
\end{itemize}

\begin{figure*}
  \centering
  \includegraphics[width=\linewidth]{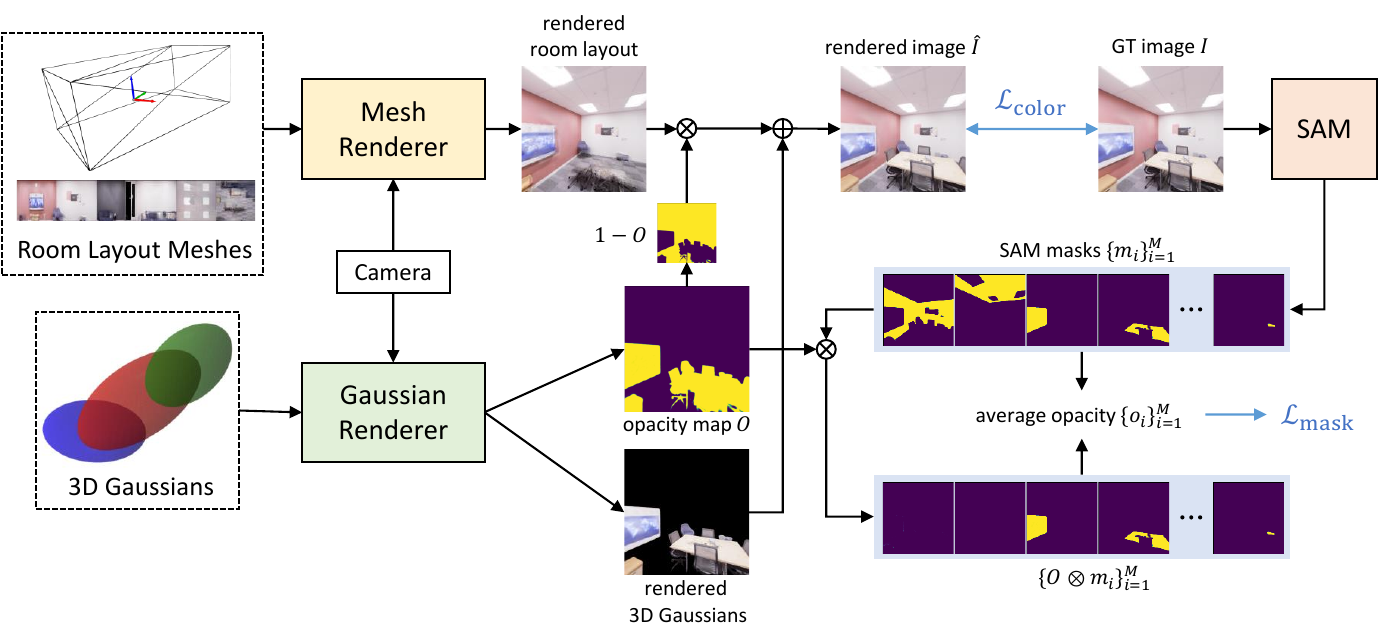}
  \caption{{\bf Overview.} Room layout meshes and 3D Gaussians are rendered using a mesh renderer and a Gaussian renderer, respectively. The outputs of the two renderers are blended to obtain the final rendered image $\hat{I}$, which is used to calculate the color loss $\mathcal{L}_\textrm{color}$ (\cref{eq:color-loss}) with the GT image $I$. Using the opacity map $O$ obtained from the Gaussian renderer and the SAM masks $\{m_i\}_{i=1}^M$ of the GT image $I$, the average opacity for each mask $\{o_i\}_{i=1}^M$ can be obtained using \cref{eq:opacity}. $\{o_i\}_{i=1}^M$ is then used to calculate $\mathcal{L}_\textrm{mask}$ as \cref{eq:mask-loss2}.}
  \label{fig:overview}
\end{figure*}

\section{Related Works}
\label{sec:related-works}
{\bf 3D Gaussian Splatting for Indoor Scenes.} 3D Gaussian Splatting (3DGS) \cite{kerbl20233d} reconstructs 3D scenes using explicit 3D Gaussians with real-time speed and high rendering quality. 360-GS \cite{bai2024360} takes four panoramic images as input to reconstruct indoor scenes using 3DGS. It initializes 3D Gaussians based on the room layout and depth maps extracted from the panoramas. FreeSplat \cite{wang2024freesplat} is designed to precisely localize 3D Gaussians from long sequence inputs. GaussianRoom \cite{xiang2024gaussianroom} integrates neural SDF with 3DGS to leverage geometric priors and further introduces monocular normal priors and edge priors to enhance the details of indoor scenes. 

{\bf Mesh Extraction from Radiance-based Method.} There have been numerous methods developed to reconstruct 3D scenes from multiple input images and extract meshes. These methods define surfaces according to the representation used for 3D scene reconstruction and then extract meshes from these surfaces. For instance, NeRF2Mesh \cite{tang2023delicate} finely extracts textured meshes through adaptive surface refinement from 3D scenes defined by NeRF \cite{mildenhall2021nerf}. NeuS \cite{wang2021neus} and NeuS2 \cite{wang2023neus2} combine NeRF and signed distance fields (SDF) to not only reconstruct 3D scenes but also estimate surfaces within continuous spaces using SDF, allowing for mesh extraction.

For 3DGS \cite{kerbl20233d}, since 3D Gaussians do not always lie on surfaces and do not directly interact with each other, the normal between neighboring Gaussians can be inconsistent. SuGaR \cite{guedon2024sugar} addresses issues by setting the opacities of 3D Gaussians to 1 and applying various regularization terms to position the 3D Gaussians on surfaces. It then samples points and uses Poisson reconstruction to extract the meshes. GSDF \cite{yu2024gsdf} and GaussianRoom \cite{xiang2024gaussianroom} integrate neural SDF within 3DGS to leverage the geometric priors of SDF and real-time speeds of Gaussian splatting. 2DGS \cite{huang20242d} and GaussianSurfels \cite{dai2024high} model surfaces by collapsing 3D Gaussians into planar Gaussian disks. DN-Splatter \cite{turkulainen2024dn} uses depth and normal priors obtained from mobile devices. 

{\bf Segment Anything Model in 3D.} Segment Anything Model (SAM) \cite{kirillov2023segment} is a 2D foundation model capable of segmenting any object in a zero-shot manner. Several works use SAM to obtain 2D segmentation masks from input images and lift them into 3D scenes. SA3D \cite{cen2023segment} and Segment Anything NeRF \cite{chen2023interactive} perform 3D segmentation and scene editing using NeRF \cite{mildenhall2021nerf} with SAM masks. SAGA \cite{hu2024semantic} and Gaussian Grouping \cite{ye2023gaussian} use SAM's 2D masks to segment 3D Gaussians and edit the reconstructed 3D scenes. Feature 3DGS \cite{zhou2024feature} and Feature Splatting \cite{qiu2024feature} distill SAM features as one of the attributes of 3D Gaussians to lift SAM's capabilities into 3D. 

\section{Method}
\label{sec:method}

\subsection{Preliminary: 3D Gaussian Splatting}

In 3D Gaussian Splatting (3DGS) \cite{kerbl20233d}, the scene is reconstructed using explicit 3D Gaussians, each characterized by following attributes: 3D position coordinates $\mu$, scale matrix $S$, rotation matrix $R$, color $c$, and its opacity $\alpha$. Each Gaussian is defined by its 3D position $\mu$ and covariance matrix $\Sigma = RSS^T R^T$ as
\begin{equation}
  G(x) = \exp (-\frac{1}{2} (x - \mu)^T \Sigma^{-1} (x - \mu) ).
  \label{eq:gs}
\end{equation}

After projecting 3D Gaussians onto the 2D image plane and sorting them by depth, rendering is achieved by alpha blending from front to back
\begin{equation}
  C = \sum_{i \in \mathcal{N}} c_i \alpha_i \prod_{j=1}^{i-1} \alpha_j .
  \label{eq:gs-color}
\end{equation}

We can optimize the 3D Gaussians by minimizing the L1 loss and D-SSIM loss between the rendered image $\hat{I}$ and the ground truth (GT) image $I$. The total loss is formulated as
\begin{equation}
  \mathcal{L}_\textrm{color} = (1 - \lambda) \mathcal{L}_1 + \lambda \mathcal{L}_\textrm{D-SSIM} ,
  \label{eq:color-loss}
\end{equation}
where $\lambda$ is a hyperparameter set to 0.2 in \cite{kerbl20233d}.

The opacity map of 3D Gaussians can be obtained during the alpha blending process as follows:
\begin{equation}
  O = \sum_{i \in \mathcal{N}} \alpha_i \prod_{j=1}^{i-1} \alpha_j .
  \label{eq:gs-opacity}
\end{equation}

\subsection{Room Layout Meshes}

3D Gaussians are effective for 3D scene reconstruction but often inefficient. For instance, when reconstructing walls inside a room, walls typically consist of planes of similar color, which can be adequately represented with a few meshes. However, in 3D Gaussian splatting, numerous Gaussians are used to represent the wall. Moreover, since those 3D Gaussians are trained independently without interaction with others, the plane fails to maintain consistent normal across the plane. This not only poses physical inaccuracies but also complicates applications such as mesh extraction. 

\begin{figure*}
  \centering
  \begin{subfigure}{0.42\linewidth}
    \includegraphics[width=\linewidth]{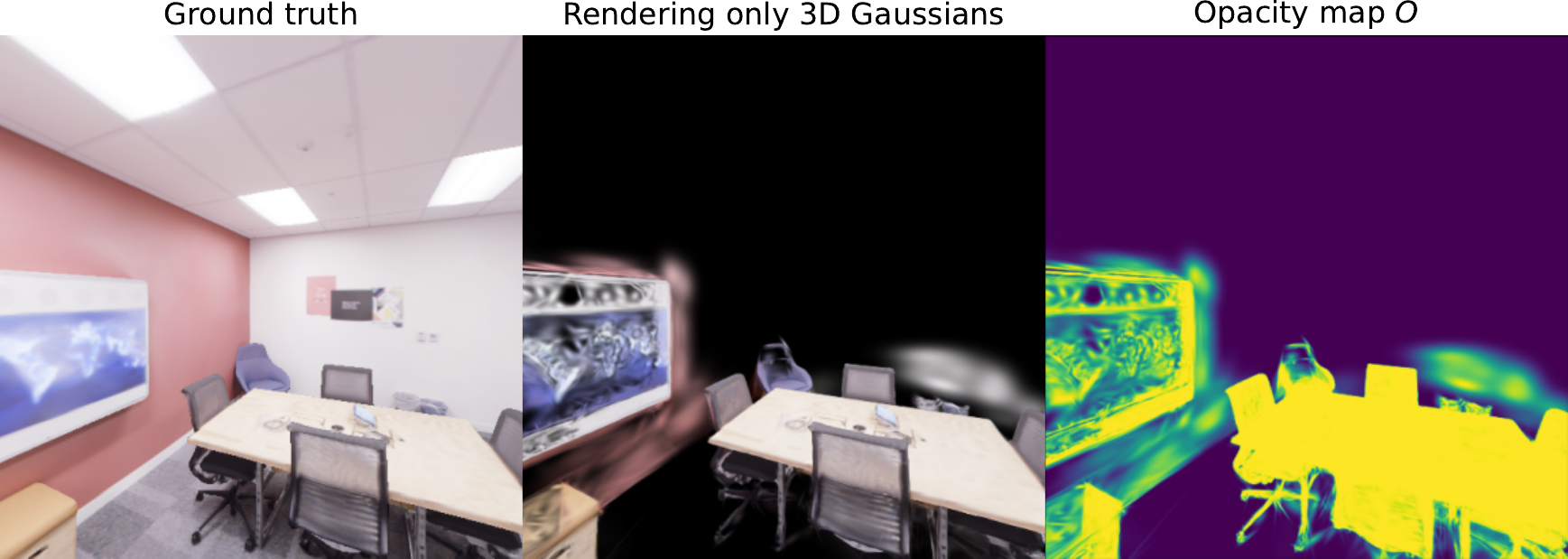}
    \caption{GT image and image rendered using only 3D Gaussians.}
    \label{fig:opacity-a}
  \end{subfigure}
  \begin{subfigure}{0.56\linewidth}
    \includegraphics[width=\linewidth]{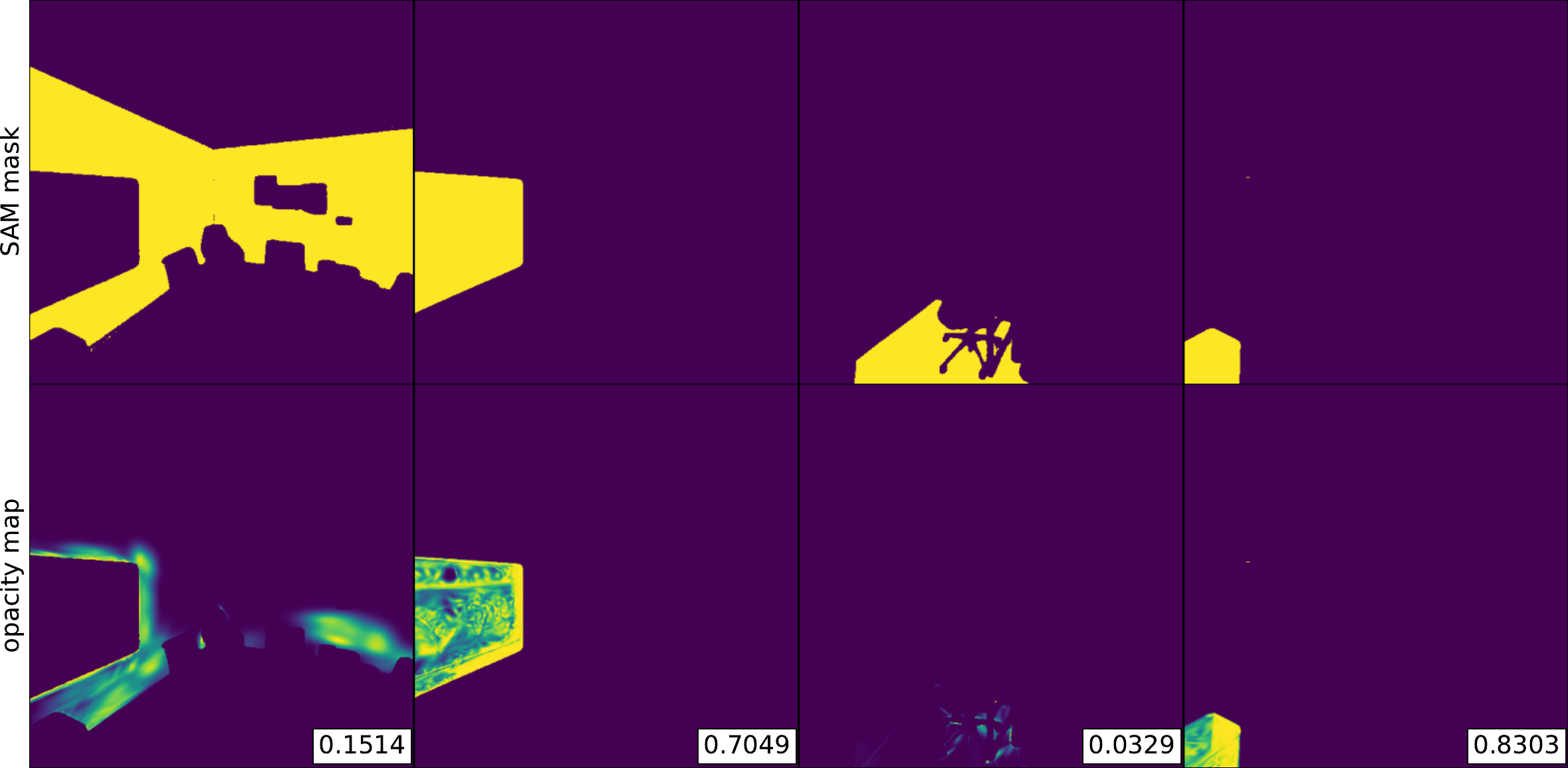}
    \caption{Opacity maps for several SAM masks.}
    \label{fig:opacity-b}
  \end{subfigure}
  \caption{Results of rendering only 3D Gaussians after naive joint training. (a) Joint training encounters optimization difficulties for the room layout and surrounding objects due to the ambiguity problem. (b) Nevertheless, the average opacity for each SAM mask serves as a suitable criterion for determining which primitive should represent each instance. The number in the bottom right corner is the average opacity for each mask.}
  \label{fig:opacity}
\end{figure*}

Therefore, we employ learnable meshes for room layouts consisting of ceilings, walls, and floors, which are rendered together with 3D Gaussians during training. Similar to 3DGS \cite{kerbl20233d}, room layout meshes exhibit subtle variations in color depending on the viewing direction due to ambient lighting and other factors. To model this, textures for each mesh utilize spherical harmonics (SH) to represent color variations accurately. Meshes using spherical harmonics are more powerful than typical RGB meshes and possess rendering capabilities comparable to 3DGS. 

The 3D vertices of the room layout meshes can be initialized with the output from a room layout estimation model \cite{sun2019horizonnet, jiang2022lgt} or approximate values obtained manually. For more accurate meshes, the coordinates of these vertices are also trained. 

3D Gaussians are initialized by Structure from Motion (e.g., COLMAP \cite{schonberger2016structure}), while the SH coefficients of room layout meshes are initialized to zero. If both primitives are jointly trained immediately after initialization, inaccurate room layout meshes can lead to incorrect optimization of 3D Gaussians. Therefore, we first pre-train room layout meshes on input images without 3D Gaussians. Then, we jointly train both primitives. 

\subsection{Joint Training of 3D Gaussians and Room Layout Meshes}

An overview of our method is shown in \cref{fig:overview}. The 3D Gaussians and room layout meshes are rendered using a differentiable mesh renderer and a differentiable Gaussian renderer, respectively. The rendering results from both renderers are blended to produce the final rendered image. 

When jointly training 3D Gaussians and room layout meshes, an ambiguity problem arises regarding which of the two primitives should be used to represent certain objects. Objects that are far from the room layout surface can be represented by 3D Gaussians based on images from multiple viewpoints without ambiguity. However, for the objects close to the room layout, it becomes difficult to distinguish whether they should be represented by the room layout meshes or by 3D Gaussians. 

Since both primitives are rendered together to match the ground truth images, the incorrect training of one can adversely affect the other. For example, if 3D Gaussians are placed right in front of the wall during the training process, they cannot be removed effectively and will persist. Moreover, if those 3D Gaussians become highly opaque, the gradient flow from the loss function to the room layout meshes behind them is reduced, preventing proper training of the room layout meshes. On the other hand, if the room layout meshes represents objects near the wall, it inhibits the formation of 3D Gaussians in front of it. This issue is particularly prevalent with texture-less objects during the training process.

A naive approach is to prune the 3D Gaussians that are close to the room layout meshes during the training process. This approach allows the room layout without nearby objects to be well-trained, as there are no unnecessary 3D Gaussians interfering. However, determining the pruning distance is challenging, as the distance to the objects can vary for each scene. Furthermore, after pruning, objects near the room layout tend to be represented predominantly by the room layout meshes due to the absence of 3D Gaussians.

\subsection{Using SAM Masks as Guidance}

The ambiguous situations mainly arise when representing objects near walls, ceilings, and floors. During training, multiple viewpoint images are used, leading to most parts of objects being represented by 3D Gaussians, while the majority of the room layout is represented by the room layout meshes. As shown in \cref{fig:opacity-a}, the objects close to the wall are represented by room layout meshes in the parts near the wall, where there are no 3D Gaussians. However, parts that are away from the wall are well represented by 3D Gaussians. Additionally, some parts of the wall itself are trained with 3D Gaussians.

One important fact is that each instance should be represented by a single primitive. Therefore, we address this limitation by leveraging the Segment Anything Model (SAM) \cite{kirillov2023segment}, a foundational model for image segmentation, to distinguish instances. We introduce an additional loss function with SAM masks to ensure that each instance is represented by a single primitive, by guiding that all pixels in the same segment mask should belong to the same primitive. Specifically, using SAM's no-prompt mode, we obtain masks for all instances. Then we can measure how much each instance is represented by 3D Gaussians in the image by applying these masks to the opacity map of 3D Gaussians obtained during training. If the average opacity for a mask is close to 1, it indicates that the instance is predominantly represented by 3D Gaussians. On the other hand, if it is close to 0, it indicates that the instance is primarily represented by room layout meshes. As shown in \cref{fig:opacity-b}, the average opacity for the mask of the room layout is close to 0, indicating it is primarily represented by room layout meshes. In contrast, the average opacity for the mask of the objects is close to 1, indicating it is predominantly represented by 3D Gaussians.

Based on this observation, we define a loss function to enforce the average opacity for each mask to be close to 0 or 1. Let the SAM mask obtained from the input image $I$ be denoted as $\{m_i\}_{i=1}^M$, where $M$ represents the total number of instance masks in the image $I$. The average opacity $o_i \in [0, 1]$ for each mask $m_i$ is defined as
\begin{equation}
  o_i = \frac{\sum_{j \in m_i} (O \otimes m_i)(j)}{\sum_{j \in m_i} m_i (j)}
  \label{eq:opacity}
\end{equation}
where $\otimes$ denotes element-wise multiplication, and $j$ is a pixel coordinate in $m_i$. $O$ is the opacity map obtained from the Gaussian rasterizer. Then, the SAM mask loss is formulated as follows:
\begin{equation}
  \mathcal{L}_\textrm{mask} = \frac{1}{M} \sum_{i=1}^M o_i (1 - o_i).
  \label{eq:mask-loss}
\end{equation}

Since $\mathcal{L}_\textrm{mask}$ is a function of opacity of 3D Gaussians, it conflicts with the densification process of standard Gaussian splatting, which periodically resets opacity of 3D Gaussians. Therefore, $\mathcal{L}_\textrm{mask}$ is applied after densification concludes, ensuring that instances are represented by a single primitive based on the trained primitives up to that point. 

\begin{figure}
  \centering
  \begin{subfigure}{0.495\linewidth}
    \includegraphics[width=\linewidth]{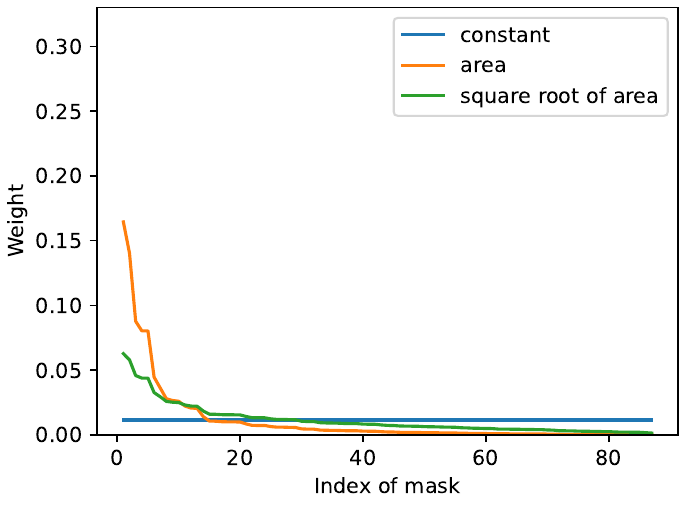}
    \caption{Allow overlapping masks.}
    \label{fig:graph-a}
  \end{subfigure}
  \begin{subfigure}{0.495\linewidth}
    \includegraphics[width=\linewidth]{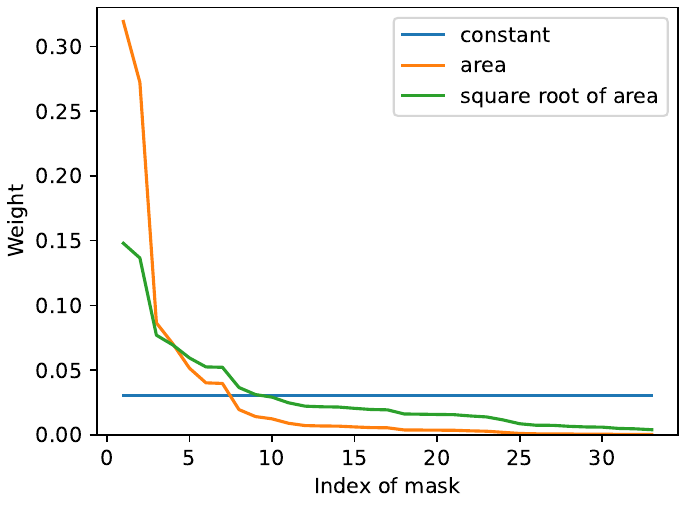}
    \caption{Remove overlapping masks.}
    \label{fig:graph-b}
  \end{subfigure}
  \caption{The weights applied to each mask based on the different weighting scheme. Applying a constant weight amplifies the influence of smaller instance masks, while using area-proportional weights amplifies larger masks more than smaller ones.}
  \label{fig:graph}
\end{figure}

Using this naive loss function presents two main issues. The first issue arises when assigning equal weights to each mask, which can amplify the influence of smaller masks. This often leads to parts of smaller objects near the room layout being represented by room layout meshes rather than 3D Gaussians. To address this, applying weights proportional to the area of each mask helps mitigate the dominance of larger masks, which otherwise could reduce the loss for smaller objects. This issue stems from the large variation in mask areas obtained from SAM, as illustrated in \cref{fig:graph}.

The second issue is that SAM segments objects at multiple levels, resulting in multiple masks for a single object. This causes the weights for each mask to diminish overall. 

To mitigate these issues, we first remove overlapping masks obtained from SAM, retaining only the largest mask. Then, we use the square root of each mask's area as its weight. The modified loss function is defined as
\begin{equation}
  \mathcal{L}_\textrm{mask} = \sum_{i=1}^M \frac{\sqrt{A_i}}{\sum_{i=1}^M \sqrt{A_i}} o_i (1 - o_i)
  \label{eq:mask-loss2}
\end{equation}
where $A_i$ is the area of mask $m_i$. The modified loss function balances the influence of masks with different areas, addressing the first issue by ensuring that both large and small masks contribute appropriately.

The final loss is a weighted sum of the original 3DGS loss $\mathcal{L}_\textrm{color}$ (\cref{eq:color-loss}) and the proposed SAM mask loss $\mathcal{L}_\textrm{mask}$:
\begin{equation}
  \mathcal{L} = \mathcal{L}_\textrm{color} + \lambda_\textrm{mask} \mathcal{L}_\textrm{mask}
  \label{eq:final-loss}
\end{equation}
where $\lambda_\textrm{mask}$ is a hyperparameter that balances the two loss terms. 

\subsection{Additional Densification}

\begin{figure*}[t]
  \centering
   \includegraphics[width=0.76\linewidth]{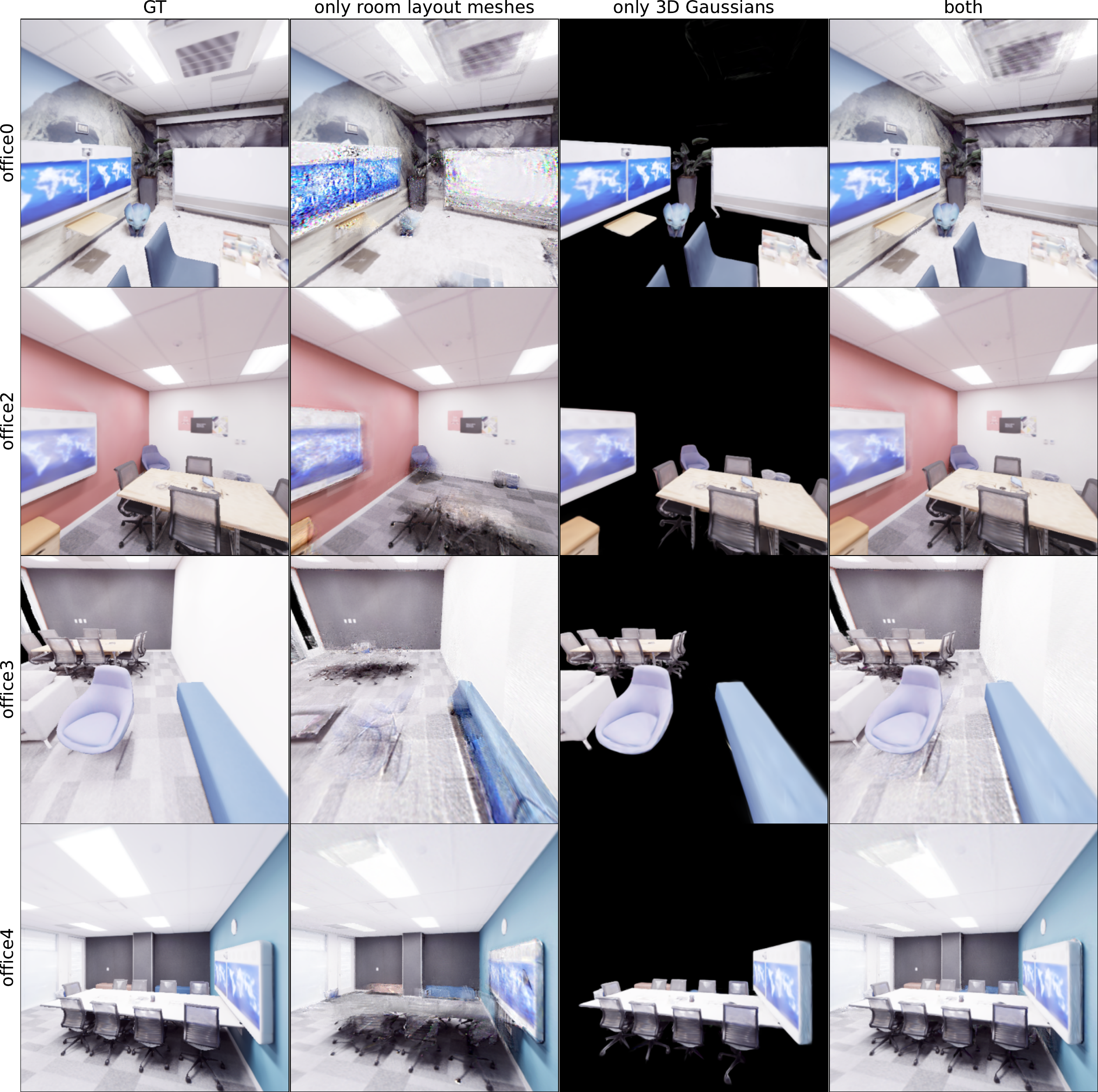}
   \caption{Qualitative results on several indoor scenes of Replica dataset \cite{straub2019replica}. The images in the second column are rendered only by the room layout meshes, and those in the third column are rendered only by the 3D Gaussians. The last column shows the combined images.}
   \label{fig:qualitative}
\end{figure*}

The SAM mask loss $\mathcal{L}_\textrm{mask}$ ensures that each instance is well represented by a single primitive. However, since $\mathcal{L}_\textrm{mask}$ is applied after standard densification has ended, it applies to a fixed number of 3D Gaussians. This leads to over-reconstruction in areas that need to be filled with a limited number of 3D Gaussians, which reduces the rendering quality of the objects as shown in \cref{fig:densification-ablation}. 

Therefore, we introduce the additional densification step to alleviate the limitation of a fixed number of Gaussians, allowing less-reconstructed objects to fill their empty areas with new Gaussians. While the standard densification resets opacity periodically, we do not reset opacity during additional densification. Instead, after splitting the target Gaussians, we set the opacity of the new Gaussians to 80\% of the opacity of the target Gaussians, ensuring that only the opacity of target Gaussians are reduced without unnecessary resets. 

\begin{table*}
  \centering
  {\small{
  \begin{tabular}{@{}l|lccccc@{}}
    \toprule
    & Method & GIoU $\uparrow$ & LIoU $\uparrow$ & PSNR $\uparrow$ & SSIM $\uparrow$ & LPIPS $\downarrow$ \\
    \midrule
    A & naive joint training & 0.7475 & 0.9333 & 30.13 & 0.9404 & 0.1210 \\
    B & A + naive SAM mask loss (\cref{eq:mask-loss}) & 0.9281 & 0.9801 & 30.38 & 0.9405 & 0.1158 \\
    C & B + remove overlapping masks & 0.9349 & 0.9829 & 30.46 & 0.9408 & 0.1151 \\
    D & C + weight proportional to the square root of the area (\cref{eq:mask-loss2}) & 0.9528 & 0.9880 & 30.60 & 0.9426 & 0.1131 \\
    Ours & D + additional densification & {\bf 0.9586} & {\bf 0.9888} & {\bf 30.66} & {\bf 0.9431} & {\bf 0.1111} \\
    \bottomrule
  \end{tabular}
  }}
  \caption{Ablation study on office 2. The SAM mask loss and additional densification both improve image quality, GIoU, and LIoU.}
  \label{tab:ablation}
\end{table*}

This approach ensures that the SAM mask loss $\mathcal{L}_\textrm{mask}$ and additional densification do not conflict, while addressing the issue where new Gaussians might be obscured by existing ones, thus enabling effective densification. 

Our additional densification is based on the split and prune operations of standard densification, but differs in that after splitting, it reduces the opacity to 80\% of the original opacity instead of copying the opacity. 

\section{Experiments}
\label{sec:experiments}

\subsection{Setup}
{\bf Datasets.} Our method is evaluated on Replica \cite{straub2019replica} dataset, which is the synthetic indoor scene dataset. Using the physics-enabled 3D simulator Habitat-Sim \cite{savva2019habitat}, various scenes from Replica dataset are simulated, and 90 images from different viewpoints are captured. One-eighth of the captured images are used as the test set, while the rest are used for training. The 3D vertices of the room layout meshes are obtained manually from the reconstructed meshes of Replica dataset. The coordinate system is transformed to have the z-axis pointing upwards.

{\bf Evaluation Metric.} In our method, we can render either 3D Gaussians and room layout meshes separately or render both primitives together. One of the key evaluation factors is whether the room layout and objects are trained well as room layout meshes and 3D Gaussians, respectively. 

Therefore, we obtain the GT semantic map for each image in the Replica dataset and use it to derive the GT binary mask $L$ for the room layout. Also we derive the GT binary mask for the remaining objects: $G = 1-L$. Then, we render only the 3D Gaussians to obtain the opacity map $O$ and calculate the Intersection over Union (IoU) with the GT binary masks. The IoU for the areas that should be represented by the 3D Gaussians and the room layout meshes are referred to as GIoU and LIoU, respectively. 
\begin{align}
  \textrm{GIoU} &= \frac{\vert G \cap O \vert}{\vert G \cup O \vert} \\
  \textrm{LIoU} &= \frac{\vert L \cap (1 - O) \vert}{\vert L \cup (1 - O) \vert}
  \label{eq:iou}
\end{align}

For rendering quality, the PSNR, SSIM, LPIPS of images rendered with both primitives together are used as evaluation metrics. 

{\bf Implementation Details.} The training of 3D Gaussians follows standard 3D Gaussian splatting \cite{kerbl20233d}. For room layout meshes, the SH coefficient is 3, and the learning rate for SH coefficients is the same as for 3D Gaussians. The texture resolution for the meshes is $512 \times 3072$. The room layout meshes are implemented and rendered using the nvdiffrast \cite{laine2020modular} library. The hyperparameter $\lambda_\textrm{mask}$ is set to 0.5. The learning rate for the 3D vertices of the meshes is set to $10^{-5}$.

The room layout meshes are pre-trained without 3D Gaussians, using a batch size of 8 for 1k iterations before joint training begins. For additional densification, it starts after standard densification is completed at the 15k iteration and continues until the end of training. 

3D Gaussians that have moved behind room layout meshes do not affect the rendered image. While it is possible to render 3D Gaussians only up to the depth rendered by the room layout meshes, those obscured by room layout meshes remain in place since they cannot receive gradient flow from the rendered image. Therefore, to save memory and simplify the code, we remove those 3D Gaussians every iteration. This process does not affect the optimization process. 

\subsection{Qualitative Result}

\begin{figure}[t]
  \centering
   \includegraphics[width=\linewidth]{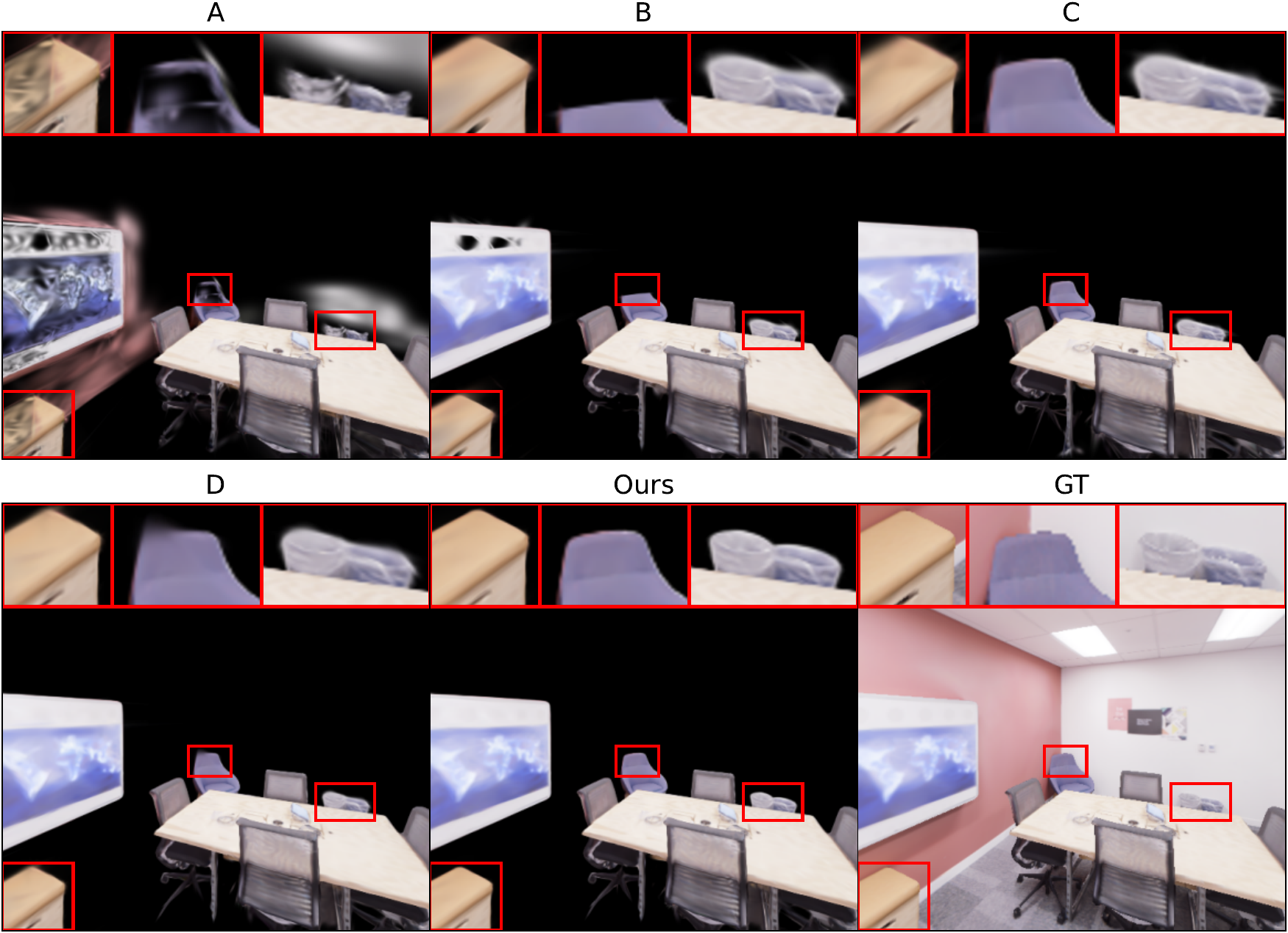}
   \caption{Ablation study on office 2. The SAM mask loss and additional densification both improve image quality and instance separation. }
   \label{fig:ablation}
\end{figure} 

We conducted experiments on various scenes from the Replica dataset \cite{straub2019replica}. The qualitative results are shown in \cref{fig:qualitative}. For all scenes, the room layout and objects are clearly represented using room layout meshes and 3D Gaussians, respectively. 

\subsection{Ablation Studies}

\begin{table*}
  \centering
  {\small{
  \begin{tabular}{@{}lcccccc@{}}
    \toprule
    Weight & \makecell{Overlapping \\ mask} & GIoU $\uparrow$ & LIoU $\uparrow$ & PSNR $\uparrow$ & SSIM $\uparrow$ & LPIPS $\downarrow$ \\
    \midrule
    \multirow{2}{*}{$\frac{1}{M}$ (\cref{eq:mask-loss})} & allow & 0.9455 & 0.9844 & 30.57 & 0.9422 & 0.1124 \\
    & remove & 0.9500 & 0.9864 & 30.63 & 0.9421 & 0.1127 \\\hline
    \multirow{2}{*}{$\frac{A_i}{\sum_{i=1}^M A_i}$} & allow & 0.8848 & 0.9872 & 30.70 & 0.9436 & {\bf 0.1099} \\
    & remove & 0.9295 & {\bf 0.9896} & {\bf 30.72} & {\bf 0.9438} & 0.1103 \\\hline
    \multirow{2}{*}{$\frac{\sqrt{A_i}}{\sum_{i=1}^M \sqrt{A_i}}$ (\cref{eq:mask-loss2})} & allow & 0.9565 & 0.9880 & 30.69 & 0.9432 & 0.1112 \\
    & remove & {\bf 0.9586} & 0.9888 & 30.66 & 0.9431 & 0.1111 \\
    \bottomrule
  \end{tabular}
  }}
  \caption{Ablation study of weighting scheme and overlapping mask on office 2. The area-proportional weighting scheme yields better image quality and LIoU compared to other weighting schemes, but significantly reduces GIoU. In contrast, the weighting scheme proportional to the square root of the area improves GIoU without greatly compromising image quality.}
  \label{tab:ablation2}
\end{table*}

\begin{figure}[t]
  \centering
   \includegraphics[width=\linewidth]{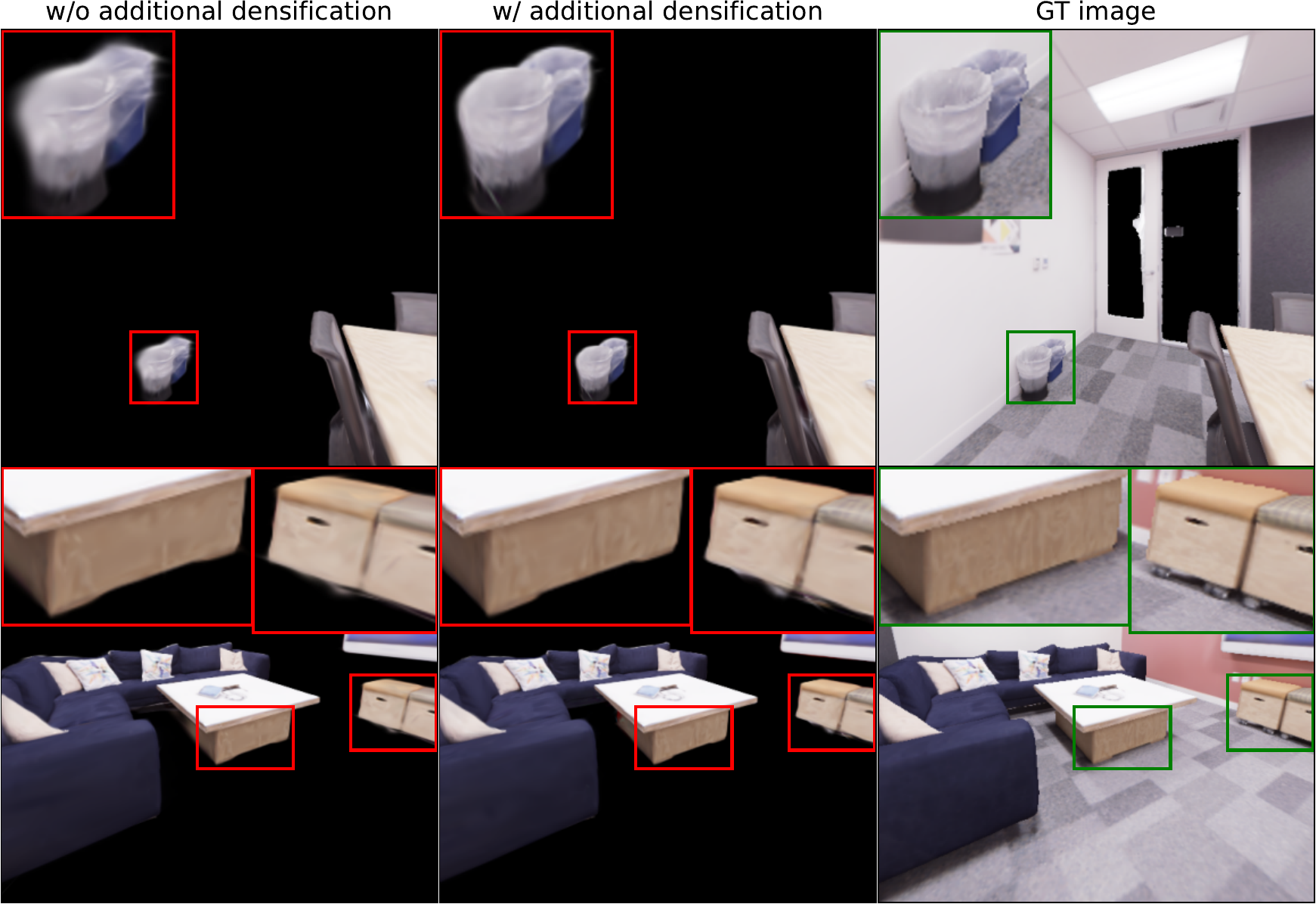}
   \caption{Ablation study of additional densification on office 2. Our additional densification improves image quality of 3D Gaussians and performance in separating primitives.}
   \label{fig:densification-ablation}
\end{figure}

We conducted an ablation study to verify the effects of the proposed SAM mask loss $\mathcal{L}_\textrm{mask}$ and additional densification. As shown in \cref{tab:ablation,fig:ablation,fig:densification-ablation}, both SAM mask loss and additional densification have positive effects on image quality and the training of appropriate primitives. Even when using a naive SAM mask loss (\cref{eq:mask-loss}), significant improvements are observed in GIoU and LIoU metrics. This indicates that using masks from SAM effectively guides the instance separation and determines which primitives should represent a 3D scene. 

We also conducted an ablation study on the weighting scheme and the use of overlapping SAM masks when applying the SAM mask loss $\mathcal{L}_\textrm{mask}$. As shown in \cref{tab:ablation2}, area-proportional weighting scheme resulted in slightly better image quality and LIoU compared to other weighting schemes. However, considering that training the appropriate primitive is as important as image quality, this scheme cannot be considered optimal because it decreases the GIoU. In contrast, using weights proportional to the square root of the area showed a slight decrease in image quality compared to the area-proportional weighting scheme, but it improves GIoU, aligning well with our goal. 

\subsection{Application: Editing Room Layout}

\begin{figure}[t]
  \centering
   \includegraphics[width=0.7\linewidth]{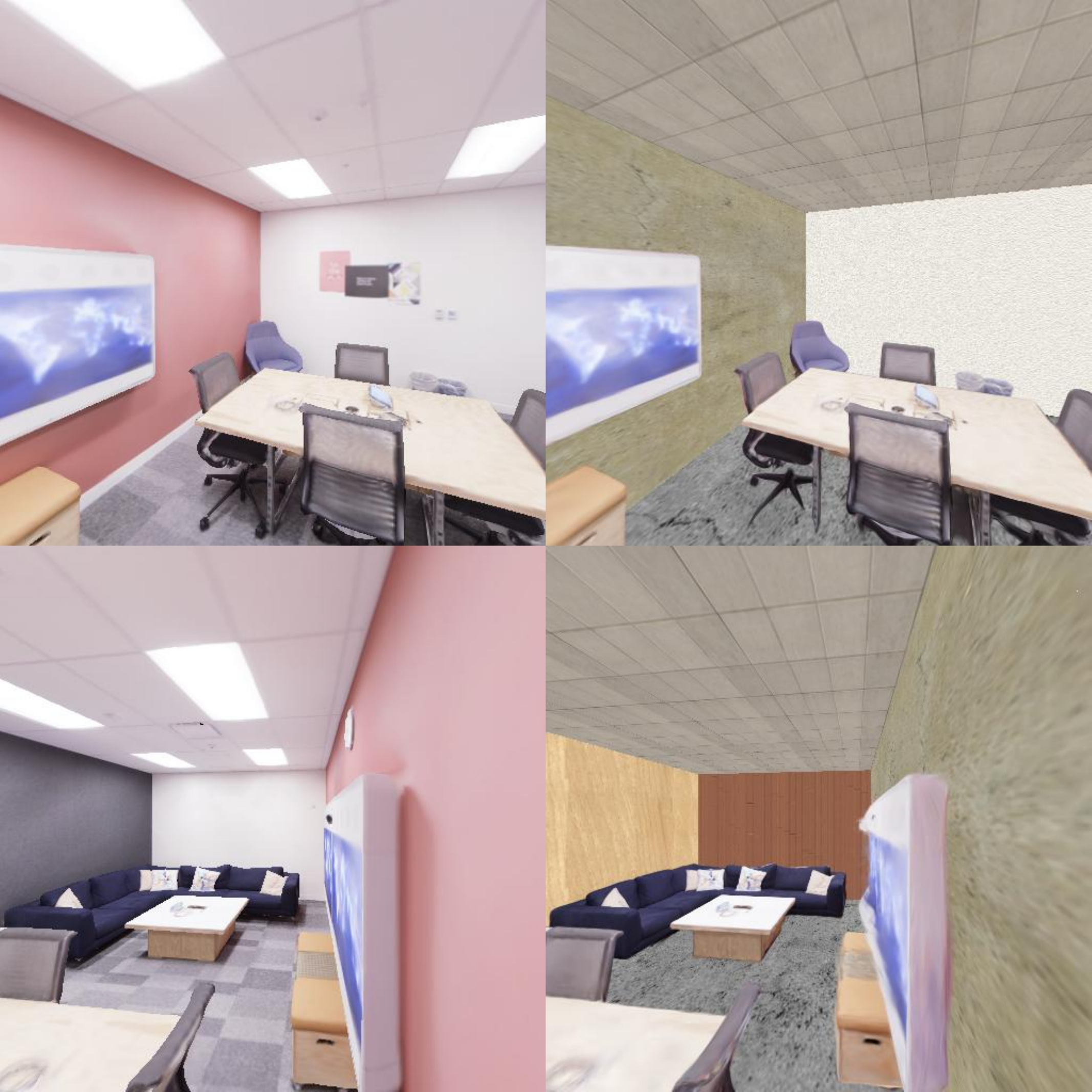}
   \caption{Examples of rendered scenes with edited room layouts. Our method uses two primitives to represent the scene, enabling room layout edits by modifying the textures of the trained room layout meshes.}
   \label{fig:application}
\end{figure}

Our method allows for the room layout to be represented by meshes and the other objects by 3D Gaussians, enabling selective use of either one. This allows for arbitrary modification or editing of one while keeping the other intact to reconstruct new scenes. Here, as an application, we demonstrate using the trained 3D Gaussians as is while changing the room layout meshes arbitrarily, with the results shown in \cref{fig:application}. The current editing pipeline is at the prototype stage, and we leave the enhancement to more realistic room layouts as future work (e.g. lighting, shadow). 

\section{Conclusion}
\label{sec:conclusion}

In this study, we reconstruct indoor scenes using room layout meshes and 3D Gaussians. To address ambiguity issues arising from jointly training these two primitives, we utilize SAM masks to represent the room layout and other objects separately as distinct primitives. We define the average opacity for each mask using SAM masks and the opacity map obtained from the 3D Gaussians. The SAM mask loss is designed to converge the average opacity for each mask to either 0 or 1. We show that appropriate weighting methods for SAM mask loss and the removal of overlapping masks can better distinguish between the two primitives during training. Moreover, we introduce additional densification without opacity reset to enhance image quality. Finally we show that the room layout of the scene can be easily modified by editing the texture maps of the trained room layout meshes. 

Our method is currently applied only to indoor scenes. However, meshes are a primitive that can be used for any scene type. In future work, we will explore the joint training of meshes and 3D Gaussians for outdoor scenes and large indoor scenes to extend our method. 

\clearpage
\section*{Acknowledgment}
This work was supported by Institute for Information \& communications Technology Promotion (IITP) grant funded by the Korea government (MSIT) [No.2021-0-01343-004, Artificial Intelligence Graduate School Program (Seoul National University)]

\clearpage
\appendix
\noindent \scalebox{1.4}{\bf Appendix}

\section{Comparison with 3D Gaussian Splatting}

\begin{table}[h]
  \centering
  {\small{
  \begin{tabular}{@{}lccccc@{}}
    \toprule
    Method & PSNR $\uparrow$ & SSIM $\uparrow$ & LPIPS $\downarrow$ & \makecell{Number of \\ 3D Gaussians $\downarrow$} \\
    \midrule
    3DGS \cite{kerbl20233d} & {\bf 32.00} & {\bf 0.9533} & {\bf 0.1066} & 429k \\
    Ours & 30.66 & 0.9431 & 0.1111 & {\bf 140k} \\
    \bottomrule
  \end{tabular}
  }}
  \caption{Comparison with 3DGS \cite{kerbl20233d} on office 2.}
  \label{tab:3dgs}
\end{table}

We also conduct a comparison between our method and 3D Gaussian Splatting (3DGS) \cite{kerbl20233d}. We evaluate our method with 3DGS based on image quality and the number of 3D Gaussians. The results are shown in \cref{tab:3dgs}.

When representing indoor scenes with two primitives, ambiguity issues often lead to degraded image quality compared to representing them solely with 3D Gaussians. While slightly compromising on image quality, our method makes editing room layouts easier compared to 3DGS by using meshes to represent the room layout. Also, our method eliminates the need to use numerous 3D Gaussians for the room layout, significantly reducing the number of 3D Gaussians required.

\section{More Results on the Ablation Study}

\begin{figure*}
  \centering
  \includegraphics[width=\linewidth]{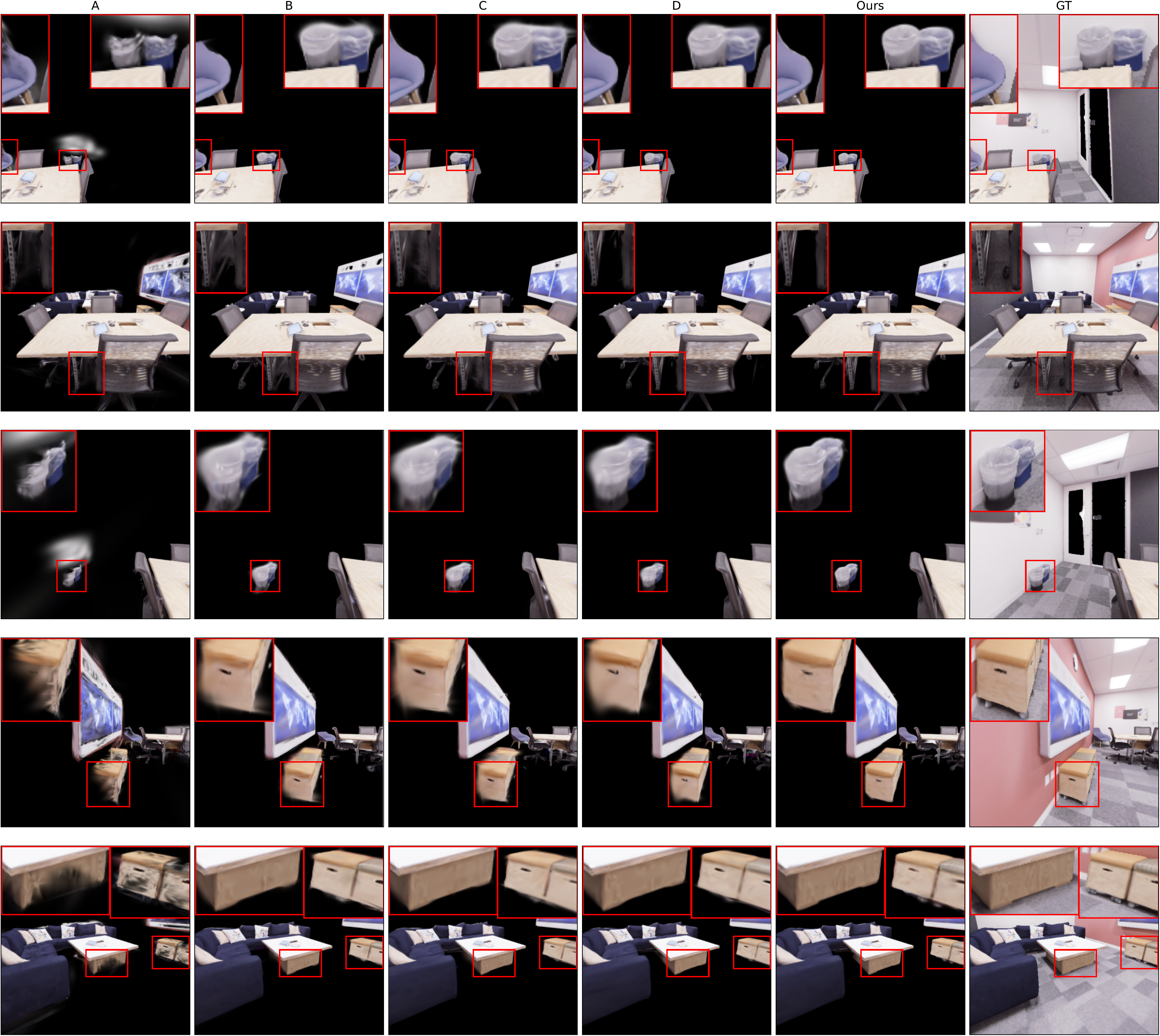}
  \caption{More results on the ablation study from various test views of office 2. Please refer to \cref{tab:ablation} for A-D. The SAM mask guides optimization to represent a single instance as one primitive. An appropriate weighting scheme enables better instance separation. Additional densification effectively improves degraded image quality.}
  \label{fig:ablation-supp}
\end{figure*}

The results of the ablation study from various test views are shown in \cref{fig:ablation-supp}.

\section{More Results on Room Layout Editing}

\begin{figure*}
  \centering
  \includegraphics[width=\linewidth]{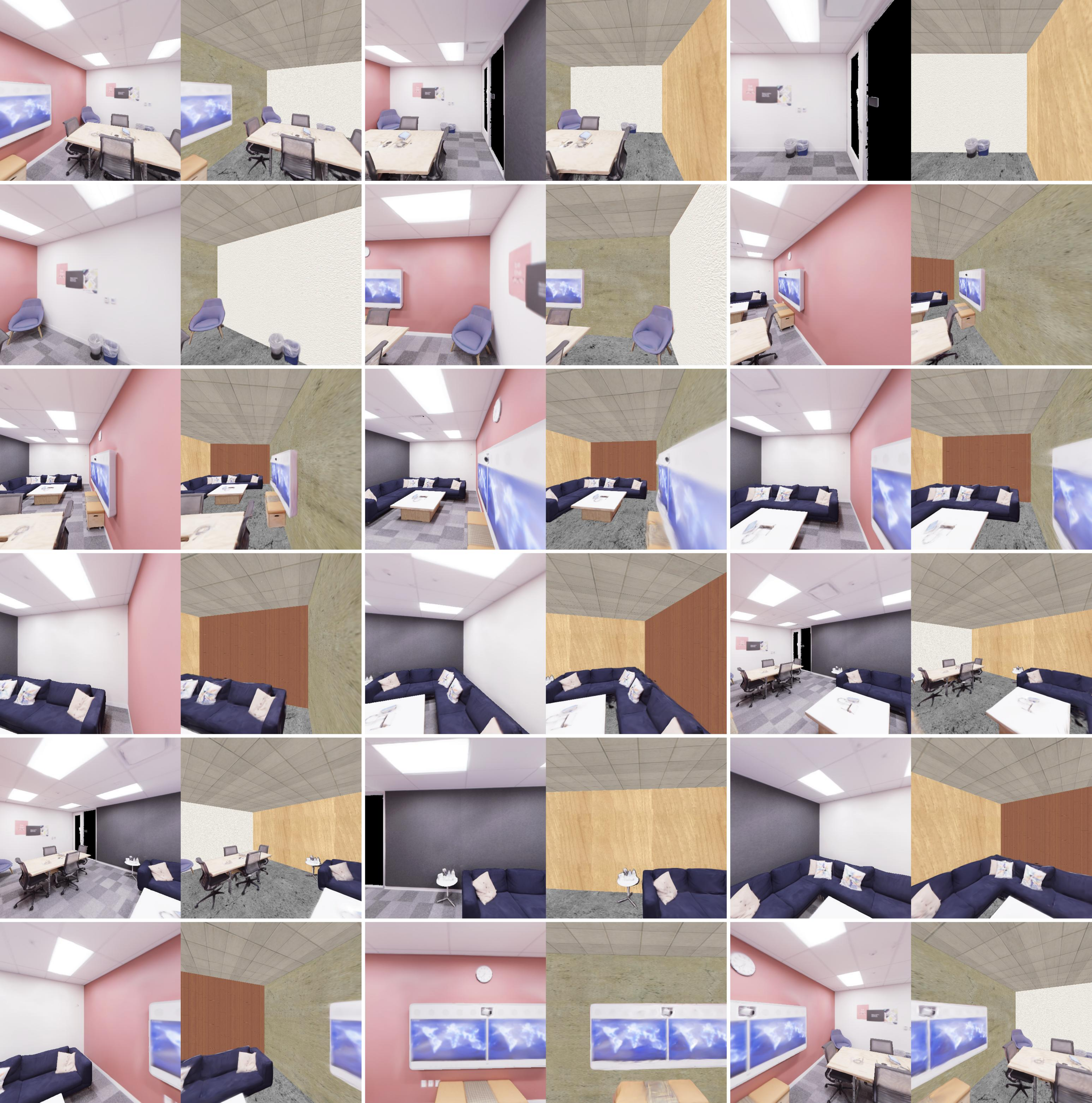}
  \caption{More results on room layout editing from various novel views. The left is the ground truth (GT) test image, and the right is the rendered image with edited room layout meshes.}
  \label{fig:application-supp}
\end{figure*}

We obtained ground truth (GT) images separately using Habitat-Sim \cite{savva2019habitat}, distinct from the train and test datasets. We then rendered scenes with edited room layouts at the viewpoints of these images. The results are shown in \cref{fig:application-supp}.

\end{document}